\documentclass[11pt]{article}
\usepackage{acl2016}
\usepackage{times}
\usepackage{latexsym}
\usepackage{url}
\usepackage{amsmath}
\usepackage{graphicx}

\aclfinalcopy 

\title{BERT for Evidence Retrieval and Claim Verification}

\author{Amir Soleimani, Christof Monz, Marcel Worring\\
	    Informatics Institute\\
	    University of Amsterdam\\
	    The Netherlands\\
	    {\tt \{a.soleimani,c.monz,m.worring\}@uva.nl}}

\date{}

\begin{document}

\maketitle

\begin{abstract}

Motivated by the promising performance of pre-trained language models, we investigate BERT in an evidence retrieval and claim verification pipeline for the FEVER fact extraction and verification challenge. To this end, we propose to use two BERT models, one for retrieving potential evidence sentences supporting or rejecting claims, and another for verifying claims based on the predicted evidence sets. To train the BERT retrieval system, we use pointwise and pairwise loss functions, and examine the effect of hard negative mining. A second BERT model is trained to classify the samples as supported, refuted, and not enough information. Our system achieves a new state of the art recall of 87.1 for retrieving top five sentences out of the FEVER documents consisting of 50K Wikipedia pages, and scores second in the official leaderboard with the FEVER score of 69.7.


\end{abstract}

\section{Introduction}

The constantly growing online textual information and the rise in the popularity of social media have been accompanied by the spread of fake news and false claims. It is not feasible to manually determine the truthfulness of such information. Therefore, there is a need for automatic verification and fact-checking. Due to the unavailability of proper datasets for evidence-based fake news detection, we focus on claim verification.

\begin{figure}[!t]
\begin{center}
\begin{tabular}{ |p{7cm}| } 
 \hline
 \textbf{Claim:} Roman Atwood is a content creator. \\ 
 \textbf{Evidence:} \textbf{\texttt{\detokenize{[wiki/Roman_Atwood]}}} He is best known for his vlogs, where he posts updates about his life on a daily basis. \\
 \textbf{Verdict:} SUPPORTED \\
 \hline
  \textbf{Claim:} Furia is adapted from a short story by Anna Politkovskaya. \\ 
 \textbf{Evidence:} \textbf{\texttt{\detokenize{[wiki/Furia_(film)]}}} Furia is a 1999 French romantic drama film directed by Alexandre Aja, who co-wrote screenplay with Grégory Levasseur, adapted from the science fiction short story Graffiti by Julio Cortázar. \\
 \textbf{Verdict:} REFUTED \\
 \hline
  \textbf{Claim:} Afghanistan is the source of the Kushan dynasty. \\ 
 \textbf{Verdict:} NOT ENOUGH INFO \\
 \hline
\end{tabular}
\end{center}
\caption{Three examples from the FEVER dataset \cite{FEVER}. Given a claim, the task is to to extract evidence sentence(s) from the Wikipedia dump and classify the claim as 'SUPPORTED', 'REFUTED', or 'NOT ENOUGH INFO'}
\label{fig_fever}
\end{figure}

The Fact Extraction and VERification (FEVER) shared task \cite{FEVER} introduces a benchmark for evidence-based claim verification. FEVER consists of 185K generated claims labelled as 'SUPPORTED', 'REFUTED' or 'NOT ENOUGH INFO' based on the introductory sections of a 50K popular Wikipedia pages dump. The benchmark task is to classify the veracity of textual claims and extract the correct evidence sentences required to support or refute the claims. The primary evaluation metric (FEVER score) is label accuracy conditioned on providing evidence sentence(s) unless the predicted label is 'NOT ENOUGH INFO', which does not need any specific evidence. Figure 1 shows three examples of the FEVER dataset.

To verify a claim, an enormous amount of information needs to be processed against the claim to retrieve the evidence and then infer possible evidence-claim relations. This problem can be alleviated by a multi-step pipeline. Most work on FEVER has adopted a three-step pipeline (Figure 2): document retrieval, sentence retrieval, and claim verification. First, a set of documents, which possibly contain relevant information to support or reject a claim, are shortlisted from the Wikipedia dump. Second, five sentences are extracted out of the retrieved documents to be used as evidence. Third, the claim is verified against the retrieved evidence sentences.

The FEVER dataset covers a wide range of topics, and to overcome data limitation pre-trained models are useful. Recently the release of Bidirectional Encoder Representations from Transformers (BERT) \cite{bert} has significantly advanced the performance in a wide variety of Natural Language Processing (NLP) tasks and datasets including MS MARCO \cite{msmarco} in passage re-ranking and MultiNLI \cite{multinli} in natural language inference that respectively resemble the second and third step of the FEVER baseline. However, to the best of our knowledge, there is no integrated work for both steps.


In this paper, we propose a three-step pipeline system to address the FEVER task. We examine the BERT model for the FEVER task, and use that for evidence retrieval and claim verification. A first BERT model is trained to retrieve evidence sentences required for verifying the claims. We compare pointwise cross entropy loss and pairwise Hinge loss and Ranknet loss \cite{ranknet} for the BERT sentence retrieval. We further investigate the effect of Hard Negative Mining (HNM). Next, we train another BERT model to verify claims against the retrieved evidence sentences. 

\begin{figure}[!ht]
\centering
   \includegraphics[width=.4\textwidth, angle=0]{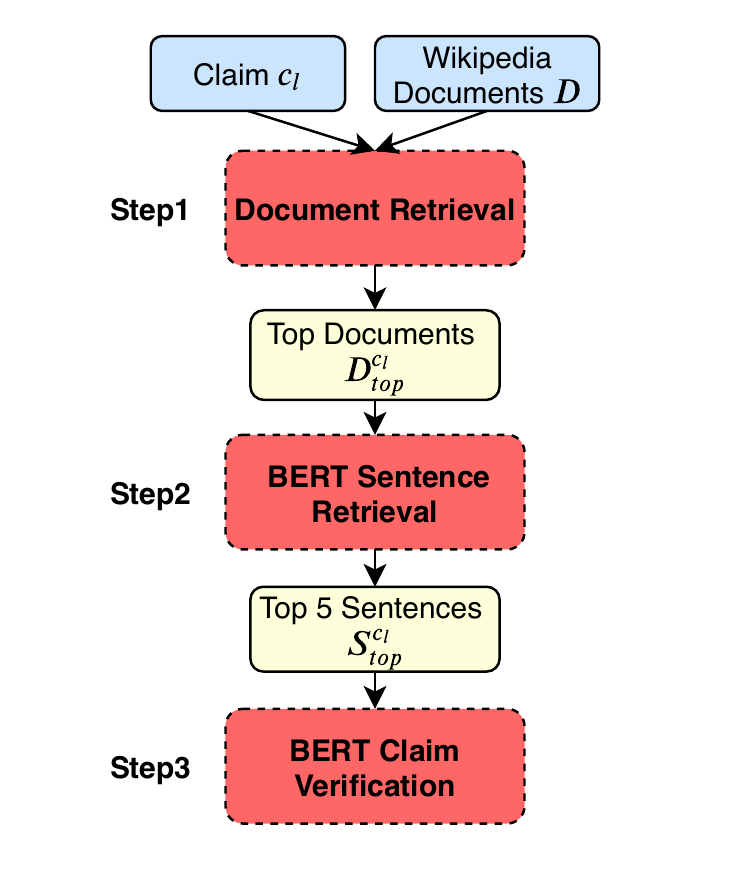}
\label{pipeline}
 \caption{Three-step pipeline evidence extraction and claim verification.}
\end{figure}


In summary, our contributions are as follows: (1) We employ the BERT model for evidence retrieval and claim verification; (2) We are the first to compare pointwise loss and pairwise loss functions for training the BERT model for sentence retrieval or fact extraction; (3) We investigate HNM to improve the retrieval performance; (4) We achieve second rank in the FEVER official leaderboard without ensembling.


\section{Related Work}

In this section, we first briefly survey related background in natural language inference. Second, we review particularly previous work on the FEVER task in the three-step pipeline: document retrieval, sentence retrieval, and claim verification

\subsection{Natural Language Inference}
Natural Language Inference (NLI) is concerned with determining if a premise entails a hypothesis.
The Stanford Natural Language Inference (SNLI) \cite{slni} and the Multi-Genre NLI (MultiNLI) corpora \cite{multinli} are the two established benchmarks for NLI. The availability of these large datasets has driven the recent advances in deep learning methods for NLI.

The deep models for NLI can be divided into two categories: (1) Models that classify the premise-hypothesis relation based on the concatenation of the premise and hypothesis fixed-size representations together with their element-wise products \cite{slni,Bowman2016}; (2) Uni-directional or bi-directional attention-based models that provide reasoning over distributional representation of the sentences \cite{rocktaschel2015reasoning,esim}.

In addition to the early improvement using context-free word representations \cite{skipgram,glove}, pre-trained language models have significantly advanced several NLP tasks. In particular, BERT \cite{bert} has achieved impressive results on several NLP tasks including NLI.

\subsection{FEVER Pipeline}

\subsubsection{Document Retrieval}
In the FEVER benchmark \cite{FEVER}, the DrQA \cite{drqa} retrieval component is considered as the baseline. They choose the k-nearest documents based on the cosine similarity of TF-IDF feature vectors. In addition to the DrQA retrieval component, the Sweeper team \cite{SWEEPer} considers lexical and syntactic features for the claim and first two sentences in the pages. The authors in \cite{papelo} use TF-IDF along with exact matching of the page titles with the claim's named entities. The UCL team \cite{ucl} highlights the pages titles, and detect them in the claims. They rank pages by logistic regression and extra features like capitalization, sentence position and token matching. Keyword matching along with page-view statistics are used in \cite{unc}. UKP-Athene \cite{athene}, the highest document retrieval scoring team, uses MediaWiki API\footnote{\url{https://www.mediawiki.org/wiki/API:Main_page}} to search the Wikipedia database for the claims noun phrases.

\subsubsection{Sentence Retrieval}

In order to extract evidence sentences, \cite{FEVER} use a TF-IDF approach similar to their document retrieval. The UCL team \cite{ucl} trains a logistic regression model on a heuristically set of features.


Enhanced Sequential Inference Model (ESIM) \cite{esim} with some small modifications has been used in \cite{unc,athene}. ESIM encodes premises and hypotheses using one Bidirectional Long Short-Term Memory (BiLSTM) with shared weights. The encoded sentences are later aligned by a bidirectional attention mechanism. The encoded and aligned sentences are combined, and another shared BiLSTM matches the two representations. Finally, a softmax layer classifies the max and mean pooled representations of the second BiLSTM. 

The UKP-Athene team \cite{athene} achieved the highest sentence retrieval recall using ESIM and pairwise training. Their model takes a claim and a pair of positive and negative sentences and predicts a similarity score for each sentence. To train the model, they use a modified Hinge loss function and a random negative sampling strategy. In other words, positive samples are trained against five randomly selected negative sentences from the top retrieved pages for each claim. Note that recall is the most important factor in this step because the FEVER score counts a prediction as true if a complete set of evidence is retrieved.

\begin{figure*}[!ht]
\centering
   \includegraphics[width=.9\textwidth, angle=0]{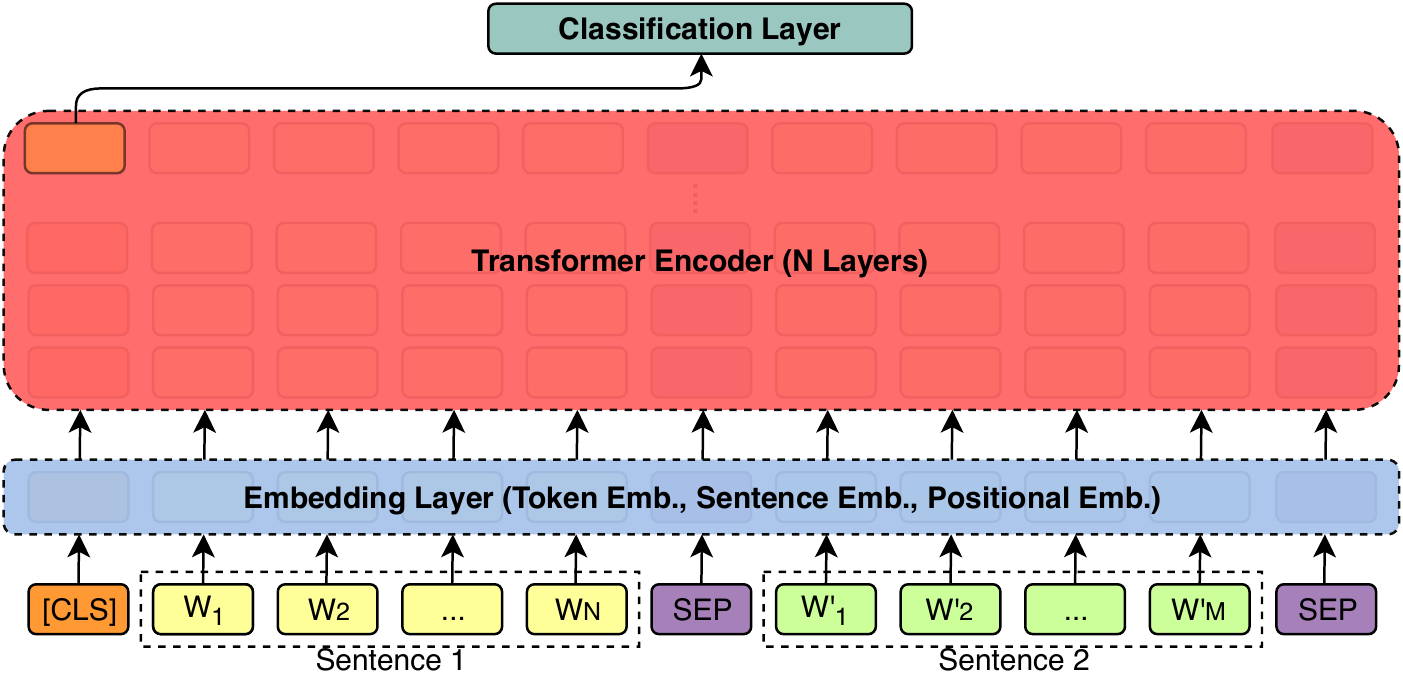}
\label{fig_bert}
 \caption{The BERT model takes the input with the form of [CLS] + Sentence 1 + [SEP] + Sentence 2 + [SEP], passes it through the embedding layer, which applies token, sentence, and positional embedding and N transformer encoder layers (BERT base:N=12, BERT large:N=24). Finally, a classification layer predicts the output from the first neuron of the last layer.}
\end{figure*}

\begin{figure*}[!ht]
\centering
   \includegraphics[width=1\textwidth, angle=0]{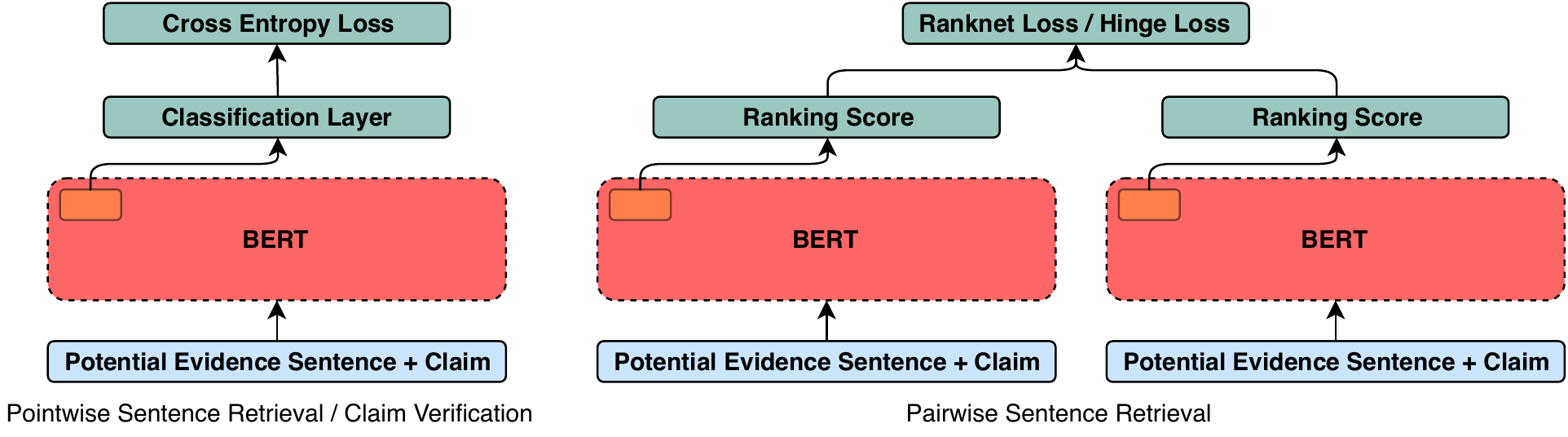}
\label{method_bert}
 \caption{Pointwise sentence retrieval and claim verification (left), Pairwise sentence retrieval (right).}
\end{figure*} 

\subsubsection{Claim Verification}
Decomposable Attention (DA) \cite{decomposableattention}, which compares and aggregates soft-aligned words in sentence pairs, is used in the FEVER benchmark paper \cite{FEVER}. The Papelo team \cite{papelo} employs transformer networks with pre-trained weights \cite{gpt}. 
ESIM has been widely used among the FEVER challenge participants \cite{unc,ucl,athene}. UNC \cite{unc}, the winner of the competition, proposes a modified ESIM that takes the concatenation of the retrieved evidence sentences and claim along with ELMo embedding and three additional token-level features: WordNet, number embedding, and semantic relatedness score from the document retrieval and sentence retrieval steps. Dream \cite{xlnetgraph} has the state of the art FEVER score. The authors use a graph reasoning method based on XLNet \cite{xlnetpaper} and RoBERTa \cite{roberta}, the two new BERT variants that are supposed to provide better pre-trained embeddings.

\section{Methods}

BERT \cite{bert} is a multi-layer transformer language representation model pre-trained on the task of next sentence prediction and masked word prediction using extremely large datasets. 

The input representation begins with a special classification embedding ([CLS]) followed by the tokens representations of the first and second sentences separated by another specific token ([SEP]). 

In order to use the BERT model for different tasks, only one additional task-specific output layer is needed that can be trained together with fine-tuning the base layers. In particular, for the classification task, a softmax layer is added on the last hidden state of the first token, which is corresponding to [CLS]. Figure 3 demonstrates the BERT model components and structures. By default we use the BERT base model (12 layers) in all the experiments.

The FEVER dataset provides $N_D$ Wikipedia documents $D=\{d_i\}^{N_D}_{i=1}$. The document $d_i$ consists of  sentences $S^{d_i}=\{s^i_j\}^{N_{S^{d_i}}}_{j=1}$. The goal is to classify the claim $c_l$ for $l=1,...,N_C$ ($N_C\!=\!145K$ for the FEVER benchmark) as 'SUPPORTED', 'REFUTED', or 'NOT ENOUGH INFO'. In order to count a prediction true, a complete set of evidence $E^{c_l}=\{s^i_j\}$ must be retrieved for the claim $c_l$. The claims with 'NOT ENOUGH INFO' label do not have an evidence set. 
\\\\
In this section, we explain the proposed system that we developed for the three FEVER steps. Figure 4 briefly demonstrates our proposed BERT-based architectures for the three-step pipeline (Figure 2).

\subsection{Document Retrieval}
In the document retrieval step, the Wikipedia documents containing the evidence supporting or refuting the claim are retrieved. Following the UKP-Athene promising document retrieval component \cite{athene}, which results in more than $93\%$ development set document recall, we exactly use their method to collect a set of top documents $D^{c_l}_{top}$ for the claim $c_l$.
\subsection{Sentence Retrieval}

The sentence retrieval step extracts the top five potential evidence sentences $S^{c_l}_{top}$ for the claim $c_l$. The training set consists of about 145K claims and all the sentences ($S^{d_i}$) from the documents retrieved at the previous step ($D^{c_l}_{top}$) corresponding to the claim $c_l$ 
($S^{c_l}_{all}=\{S^{d_i} | {d_i}\in D^{c_l}_{top}\}$). Note that $S^{c_l}_{all}$ may or may not contain the actual evidence sentences that we know from the ground-truth labels.

We adopt the pre-trained BERT model and fine-tune using two different pointwise and pairwise approaches. We did not observe any improvement to use the large BERT for this step. In both approaches, the input consist of a potential evidence sentence from $S^{c_l}_{all}$ and a claim $c_l$. Similar to \cite{papelo}, in order to compensate for the missed co-reference pronouns in the sentences, we add the Wikipedia pages titles at the beginning of each potential sentence. For all the retrieval experiments, we adopt a batch size of 32, a learning rate of $2e\!-\!5$, and one epoch of training.

\subsubsection{Pointwise}
In the pointwise approach, every single input is classified as evidence or non-evidence. We use cross entropy classification loss for the pointwise approach:
\begin{equation}
Loss_{point}=-\sum^{N}_{i=1} y_i\log(p_i)
\end{equation}
where $y_i$ and $p_i$ are respectively the one-hot ground-truth label vector and the corresponding softmax output (Figure 4 (left)), and $N$ is the total number of training samples.

At testing time, sentences are sorted by their $p_i$ values and the top five sentences are considered as evidence. A threshold can also be used on the output scores to filter out uncertain results and trade-off the recall against the precision.

\begin{table*}[t!]
\begin{center}
\begin{tabular}{|l|ccc|}
\hline \textbf{Model} & \textbf{Precision(\%)} & \textbf{Recall@5(\%)} & \textbf{F1(\%)} \\ \hline
UNC \cite{unc} & 36.39 & 86.79 & 51.38 \\
UCL \cite{ucl} & 22.74** & 84.54 & 35.84 \\
UKP-Athene \cite{athene} & 23.67* & 85.81* & 37.11* \\
DREAM-XLNet \cite{xlnetgraph} & 26.60 & 87.33 & 40.79 \\
DREAM-RoBERTa \cite{xlnetgraph} & 26.67 & 87.64 & 40.90 \\
\hline
Pointwise & 25.14 & 88.25 & 39.13 \\
Pointwise + Threshold & \textbf{38.18} & 88.00 & \textbf{53.25} \\ 
Pointwise + HNM & 25.13 & 88.29 & 39.13 \\
\hline
Pairwise Ranknet & 24.97 & 88.20 & 38.93 \\
Pairwise Ranknet + HNM & 24.97 & \textbf{88.32} & 38.93 \\
\hline
Pairwise Hinge & 24.94 & 88.07 & 38.88 \\
Pairwise Hinge + HNM & 25.01 & 88.28 & 38.98 \\
\hline
\end{tabular}
\end{center}
\caption{\label{table1} Development set sentence retrieval performance. * We calculated the scores using the official code, and for ** we used the F1 formula to calculate the score.}
\end{table*}

\subsubsection{Pairwise}
In the pairwise approach, a pair of positive and negative samples are compared against each other (Figure 4 (right)). We use the Ranknet loss function \cite{ranknet}:
\begin{equation}
Loss^{Ranknet}_{pair}=-\sum^{N}_{i=1}\log p'_i
\end{equation}
where the mapping from the positive sample $o_{pos}$ and negative sample output $o_{neg}$ to probabilities are calculated using the softmax function $p'_i=e^{o_{pos}-o_{neg}}/(1+e^{o_{pos}-o_{neg}})$. Note that we do not force the positive and negative samples to be selected from the same claims because the number of sentences per claim is significantly different and this difference might result in biasing on the claims with higher number of sentences. 

In addition, we experiment with the modified Hinge loss functions like \cite{athene}:
\begin{equation}
Loss^{Hinge}_{pair}=\sum^{N}_{i=1}\max(0,1+o_{neg}-o_{pos})
\end{equation}

At testing time, for both pairwise loss functions, we sort the sentences by their output value $o$ and similarly choose $S^{c_l}_{top}$ for the claim $c_l$.

\begin{table*}[t!]
\begin{center}
\begin{tabular}{|l|cc|}
\hline \textbf{Model} & \textbf{FEVER Score(\%)} & \textbf{Label Accuracy(\%)} \\ \hline
UNC \cite{unc} & 66.14 & 69.60 \\
UCL \cite{ucl} & 65.41 & 69.66 \\
UKP-Athene \cite{athene} & 64.74 & - \\
\hline
BERT \& UKP-Athene & 69.79 & 71.70 \\
BERT Large \& UKP-Athene & 70.64 & 72.72 \\
\hline
BERT \& BERT (Pointwise) & 71.38 & 73.51  \\
BERT \& BERT (Pointwise + HNM) & 71.33 & 73.54 \\
BERT (Large) \& BERT (Pointwise) & \textbf{72.42} & 74.58 \\
BERT (Large) \& BERT (Pointwise + HNM) & \textbf{72.42} & \textbf{74.59} \\
\hline

BERT \& BERT (Pairwise Ranknet) & 71.02 & 73.22 \\
BERT \& BERT (Pairwise Ranknet + HNM) & 70.99 & 73.02 \\
\hline
BERT \& BERT (Pairwise Hinge) & 71.60 & 72.74 \\
BERT \& BERT (Pairwise Hinge + HNM) & 70.70 & 72.76 \\
\hline
\end{tabular}
\end{center}
\caption{\label{table2} Development set verification scores. }
\end{table*}

\subsubsection{Hard Negative Mining}

The ratio of negative (non-evidence) to positive (evidence) sentences is high, thus it is not reasonable to train on all the negative samples. Random sampling limits the number of negative samples, however, this might lead to training on easy and trivial samples. Therefore, we opt to investigate the effect of HNM.

Similar to \cite{triplet}, we focus on online HNM. We fix the positive samples batch size of 16 but heuristically increase negative sample batch size from 16 to 64 and train on the positive samples and only the 16 negative samples with the highest loss values. This results in a balanced batch sized of 32. In the case of pairwise retrieval, HNM is applied to select the 32 hardest pairs out of 128 pairs thus plenty of the positive samples might not be trained on. Therefore, for this case, we heuristically increase the training epoch from one to three. While increasing the epoch for the HNM improves the performance, we observed the reverse for the normal training. In the experiments without HNM, we use random negative sampling.
Note that for both cases, loss values are computed in the no-gradient mode, like the inference time, and thus there is no need for more GPUs than normal training with the batch size of 32.

\subsection{Claim Verification}
In the final step, the top five potential evidence sentences $S^{c_l}_{top}$ are independently compared against the claim $c_l$ and the final label is determined by aggregating the five individual decisions. Like \cite{papelo}, the default label is 'NOT ENOUGH INFO' unless there is any supporting evidence to predict the claim label as 'SUPPORTED'. If there is at least one piece of evidence rejecting the claim while there is no supporting fact, the final decision is 'REFUTED'. 

We propose to train a new pre-trained BERT model as a three-class classifier (Figure 4 (left)). We train the model on $722K$ evidence-claim pairs provided by the first two steps. We adopt the batch size of $32$, the learning rate of $2e\!-\!5$, and two epochs of training.

\section{Results}

\begin{figure}[t]
\centering
   \includegraphics[width=.5\textwidth, angle=0]{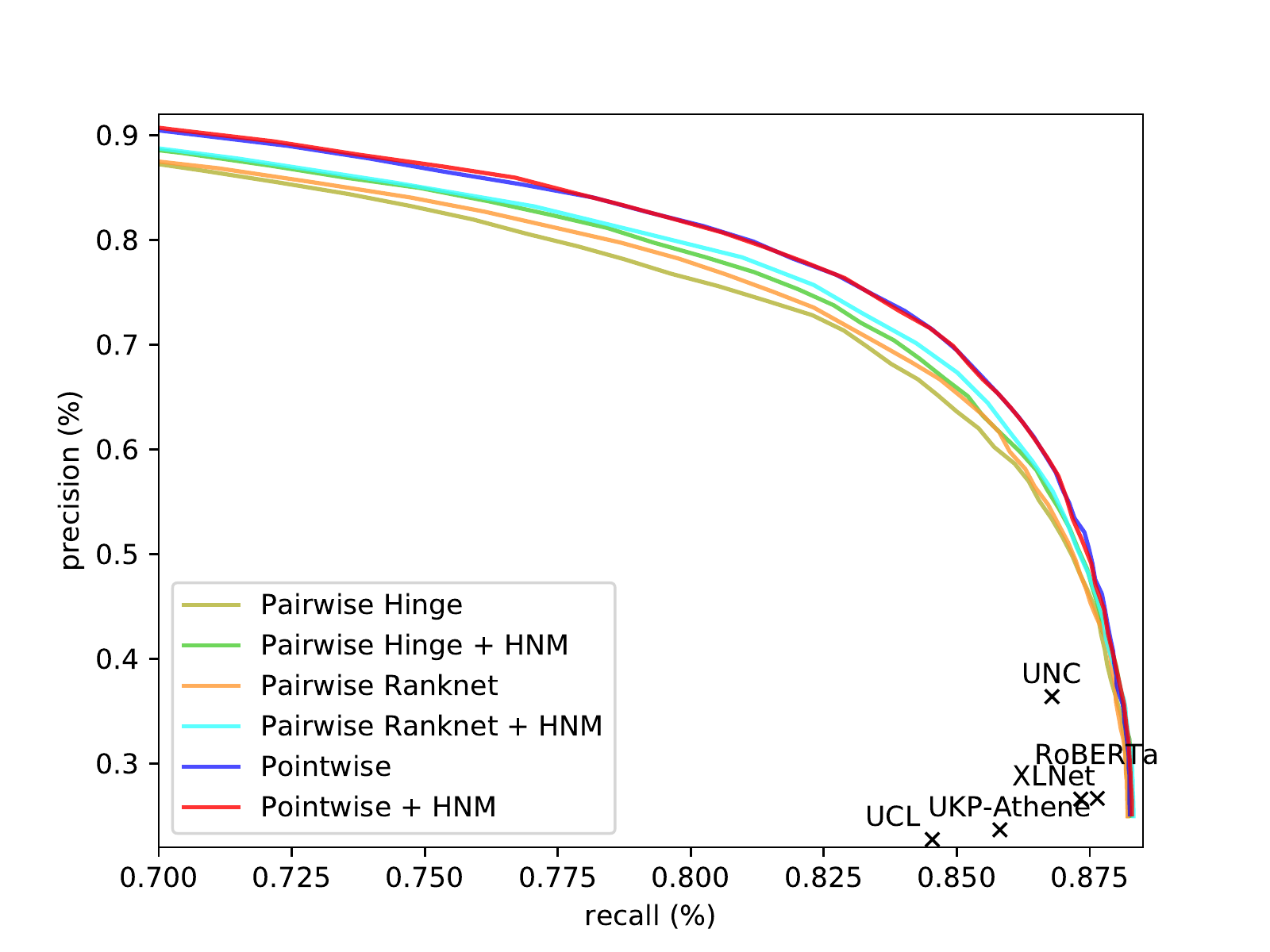}
\label{recall_precision}
 \caption{Recall and precision results on the development set. x shows the UNC, UCL, UPK-Athene, DREAM XLNet, and DREAM RoBERTa scores \cite{unc,ucl,athene,xlnetgraph}}
\end{figure}

\begin{table*}[!t]
\begin{center}
\begin{tabular}{|l|cc|}
\hline \textbf{Model} & \textbf{FEVER Score(\%)} & \textbf{Label Accuracy(\%)} \\ \hline
DREAM \cite{xlnetgraph} & \textbf{70.60} & \textbf{76.85} \\
BERT (Large) \& BERT (Pointwise + HNM) & {69.66} & {71.86} \\
abcd\_zh* & 69.40 & 72.81 \\
BERT (Large) \& BERT (Pointwise) & 69.35 & 71.48  \\
cunlp* & 68.80 & 72.47 \\
BERT \& BERT (Pointwise) & 68.50 & 70.67  \\
BERT (Large) \& UKP-Athene & 68.36 & 70.41  \\
BERT \& FEVER UKP-Athene & 67.49 & 69.40  \\
UNC \cite{unc} & 64.21 & 68.21 \\
UCL \cite{ucl} & 62.52 & 67.62 \\
UKP-Athene \cite{athene} & 61.58 & 65.46 \\
\hline
\end{tabular}
\end{center}
\caption{\label{table3} Results on the test set as of the date of writing (September 2019). * Unpublished methods listed on the leaderboard on codalab.}
\end{table*}


Table 1 compares the development set performance of different variants of the proposed sentence retrieval method with the state of the art results on the FEVER dataset. The results indicate that both pointwise and pairwise BERT sentence retrieval improve the recall. The UNC and DREAM precision scores are better than our methods without a decision threshold, however, a threshold can regulate the trade-off between the recall and precision, and achieve the best precision and F1 scores. As discussed in \cite{unc}, the recall is the most important factor. It is because the sentence retrieval predictions are the samples that we train the verification system on. Moreover, the FEVER score requires evidence for 'SUPPORTED' and 'REFUTED' claims. Therefore, we opt to focus more on recall and train the claim verification model on the predictions with the maximum recall. Surprisingly, the DREAM paper \cite{xlnetgraph} reports lower recalls for RoBERTa and XLNet that might be because of different training setups.

Although the pairwise Ranknet with HNM has the best recall score, we cannot conclude that pairwise methods are necessarily better for this task. This is more clear in Figure 5, which plots the recall-precision trade-off by applying a decision threshold on the output scores. The pointwise methods surpass the pairwise methods in terms of recall-precision performance. Figure 5 also shows that HNM enhances both pairwise methods trained by the Ranknet and Hinge loss functions and preserves the pointwise performance.

In Table 2, we compare the development set results of the state of the art methods with the BERT model trained on different retrieved evidence sets. The BERT claim verification system even if it is trained on the UKP-Athene sentence retrieval component \cite{athene}, the state of the art method with the highest recall, improves both label accuracy and FEVER score. Training based on the BERT sentence retrieval predictions significantly enhances the verification results because while it explicitly improves the FEVER score by providing more correct evidence sentences, it provides a better training set for the verification system. The large BERTs are only trained on the best retrieval systems, and as expected significantly improve the performance.

Finally, we report the blind test set results in Table 3 using the official FEVER framework on CodaLab\footnote{\url{https://competitions.codalab.org/competitions/18814#results}} as of the date of writing. Our best model ranks at the second place that indicates the importance of using pre-trained language modelling methods for both sentence retrieval and claim verification systems. Note that it is not completely fair to compare our method with the DREAM's core idea because in addition to a graph-based reasoning approach they use XLNet, a superior pre-trained language model.
\section{Conclusion}
We investigated the BERT model for evidence sentence retrieval and claim verification. In the retrieval step, we compared the pointwise and pairwise approaches and concluded that although the pairwise Ranknet approach achieved the highest recall, pairwise approaches are not necessarily superior to the pointwise approach particularly if precision is taken into account. Our large system scored second with a FEVER score of 69.66 without ensembling. 

We additionally examined hard negative mining for training the retrieval systems and showed that it slightly improves the performance. We discussed that by constantly switching between the training and inference mode, the online hard negative mining does not require additional GPUs. We leave its probable effect on the faster training to future work. Furthermore, using BERT as an end-to-end framework for the entire FEVER pipeline can be investigated in the future.

\section*{Acknowledgments}
This research was partly supported by VIVAT.
\bibliography{acl2016.bib}
\bibliographystyle{acl2016}

\end{document}